\documentclass{article}

\usepackage{microtype}
\usepackage{amsmath}
\usepackage{amssymb}
\usepackage{mathtools}
\usepackage{pgfplots}
\usepackage{amsthm, bm}
\usepackage{graphicx}
\usepackage{subfigure}
\usepackage{multirow} 
\usepackage{booktabs} 
\usepackage{nicematrix}
\usepackage{listings}
\definecolor{codegreen}{rgb}{0,0.6,0}
\definecolor{codegray}{rgb}{0.5,0.5,0.5}
\definecolor{codepink}{RGB}{252, 142, 172}
\definecolor{codepurple}{rgb}{0.58,0,0.82}
\definecolor{backcolour}{RGB}{245,245,245}
\lstdefinestyle{mystyle}{
    backgroundcolor=\color{backcolour},   
    commentstyle=\color{magenta},
    keywordstyle=\color{blue},
    numberstyle=\tiny\color{codegray},
    stringstyle=\color{codepurple},
    basicstyle=\fontfamily{\ttdefault}\footnotesize,
    breakatwhitespace=false,         
    breaklines=true,                 
    keepspaces=true,    
    frame=single,
    numbersep=5pt,                  
    showspaces=false,                
    showstringspaces=false,
    showtabs=false,                  
    tabsize=2,
    classoffset=1, %
    keywordstyle=\color{violet},
    classoffset=0,
}
\usepackage{booktabs} 

\usepackage{hyperref}

\usepackage[accepted]{icml2024}

\usepackage{amsmath}
\usepackage{amssymb}
\usepackage{mathtools}
\usepackage{amsthm}
\usepackage{pifont}

\usepackage[capitalize,noabbrev]{cleveref}

\DeclareMathOperator*{\argmax}{arg\,max}
\usepackage{array}
\usepackage{multirow}
\usepackage{booktabs}
\usepackage{float}

\usepackage{caption}
\usepackage{longtable}
\definecolor{good}{rgb}{0.11, 0.77, 0.11}
\definecolor{bad}{rgb}{0.77, 0.11, 0.11}

\usepackage{algorithm}
\usepackage{algorithmic}

\theoremstyle{plain}

\theoremstyle{definition}

\theoremstyle{remark}

\usepackage[textsize=tiny]{todonotes}

\icmltitlerunning{LLMs as Visual Explainers: Advancing Image Classification with Evolving Visual Descriptions}

\begin{document}

\twocolumn[
\icmltitle{LLMs as Visual Explainers: Advancing Image Classification with \\ Evolving Visual Descriptions}



\icmlsetsymbol{equal}{*}

\begin{icmlauthorlist}
\icmlauthor{Songhao Han}{equal,bh}
\icmlauthor{Le Zhuo}{equal,bh}
\icmlauthor{Yue Liao}{bh}
\icmlauthor{Si Liu}{bh}
\end{icmlauthorlist}

\icmlaffiliation{bh}{Beihang University, Beijing, China}

\icmlcorrespondingauthor{Yue Liao}{liaoyue.ai@gmail.com}

\icmlkeywords{Machine Learning, ICML}

\vskip 0.3in
]



\printAffiliationsAndNotice{\icmlEqualContribution} 

\begin{abstract}
Vision-language models (VLMs) offer a promising paradigm for image classification by comparing the similarity between images and class embeddings. A critical challenge lies in crafting precise textual representations for class names. While previous studies have leveraged recent advancements in large language models (LLMs) to enhance these descriptors, their outputs often suffer from ambiguity and inaccuracy. We attribute this to two primary factors: 1) the reliance on single-turn textual interactions with LLMs, leading to a mismatch between generated text and visual concepts for VLMs; 2) the oversight of the inter-class relationships, resulting in descriptors that fail to differentiate similar classes effectively. In this paper, we propose a novel framework that integrates LLMs and VLMs to find the optimal class descriptors. Our training-free approach develops an LLM-based agent with an evolutionary optimization strategy to iteratively refine class descriptors. We demonstrate our optimized descriptors are of high quality which effectively improves classification accuracy on a wide range of benchmarks. Additionally, these descriptors offer explainable and robust features, boosting performance across various backbone models and complementing fine-tuning-based methods.
\end{abstract}

\section{Introduction}
\label{sec:intro}
In recent years, a plethora of vision-language models~(VLMs)~\cite{CLIP,flamingo,blip2} has emerged, showcasing impressive transfer learning capabilities across diverse visual tasks. These models, by pretraining on large datasets, learn to align images and text within a shared embedding space. Unlike conventional models, VLMs classify images by computing the similarity between the input image and textual descriptions. Notably, CLIP~\cite{CLIP} achieves outstanding results in zero-shot image classification tasks across various datasets. This is achieved by employing a combination of class names and pre-defined templates as input prompts and then matching images to the most similar prompt.

\begin{figure}[t]
   \centering
   \includegraphics[width=\linewidth]{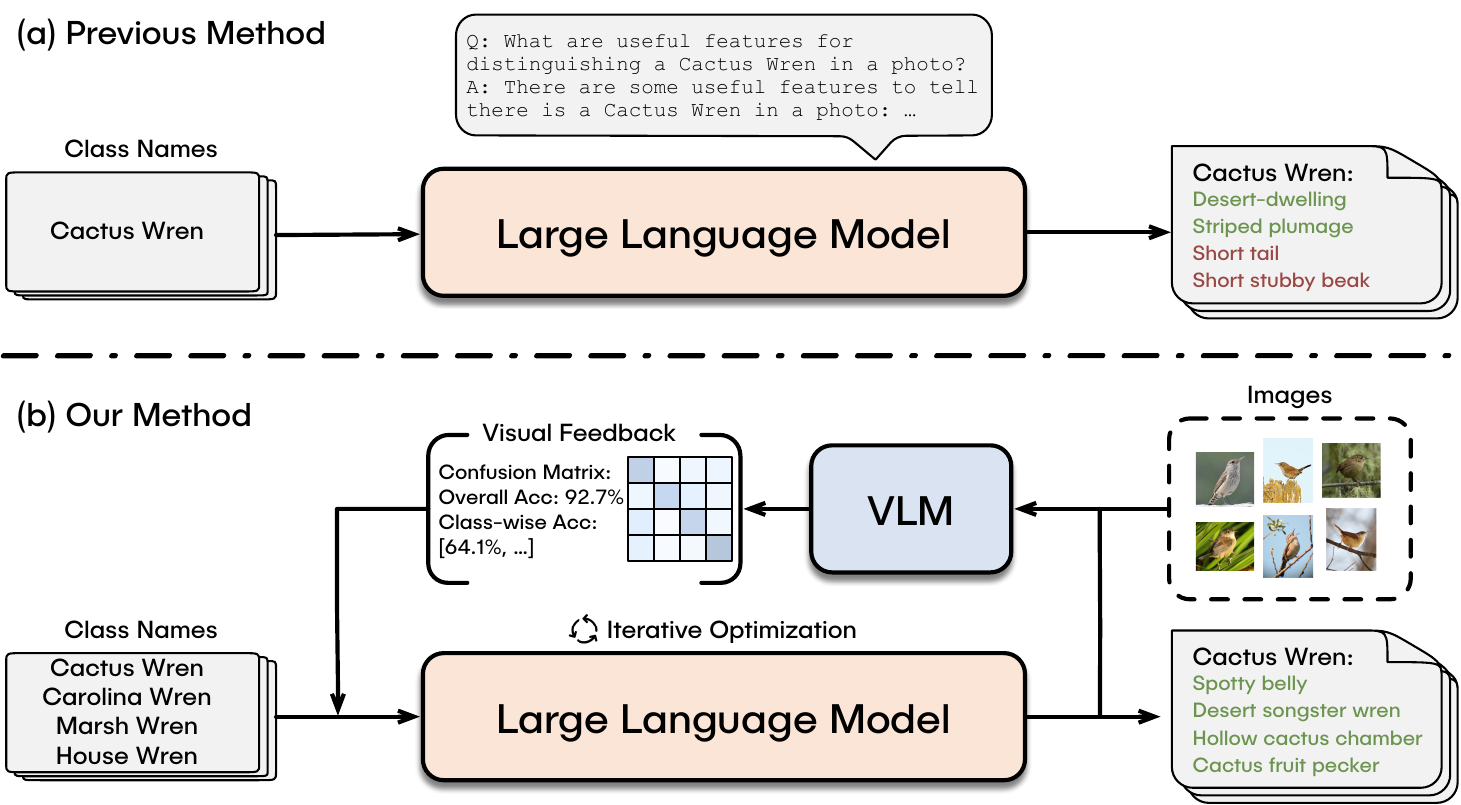} 
   \vspace{-2mm}
   \caption{\textbf{Schematic of the method.} (a) Previous methods use an LLM to generate descriptive prompts for each class directly. (b) Our method optimizes class descriptions through an evolutionary process. We utilize a VLM (such as CLIP~\cite{CLIP}) to obtain visual feedback,~\emph{e.g.}, the confusion matrix, assessing the quality of current descriptions. Upon building the visual feedback, an LLM generates refined category descriptions, iterating multiple times to achieve the final optimal category descriptions.}
   \vspace{-2mm}
   \label{fig1}
\end{figure}
 
This paradigm, though effective, is highly dependent on the quality of class prompts. For instance, in datasets with abstract or ambiguous class names, such as CUB~\cite{CUB} and Flowers102~\cite{flowers}, CLIP struggles to effectively distinguish images using class names as the sole prompts. The advent of Large Language Models (LLMs)~\cite{gpt3,gpt4,palm} has prompted research~\cite{DCLIP, WaffleCLIP, CHiLS, CuPL} into enhancing class descriptions through LLMs. These methods exploit LLMs' extensive world knowledge to generate more detailed and semantically rich descriptions for each category, thereby enriching the class prompts. Despite these advancements, current approaches exhibit several drawbacks. LLMs, trained exclusively on text, lack a nuanced understanding of visual concepts. Consequently, when provided only with textual class names, LLMs tend to produce ambiguous or inaccurate descriptions~\cite{DCLIP},~\emph{e.g.},~\emph{“short stubby beak"} for Cactus Wren, which actually has curved and relatively long beaks. Evidence from WaffleCLIP~\cite{WaffleCLIP} also suggests replacing LLM-generated class descriptions with random, meaningless characters does not hurt the overall classification performance, questioning the effectiveness of these methods. Moreover, the fundamental goal of LLM-based methods is to approximate the global optimal centroids within the CLIP embedding space for all classes using descriptive texts generated by LLMs. Achieving this necessitates considering inter-category relationships and engaging in an iterative optimization process. Current methodologies, as depicted in Figure~\ref{fig1}, are tailored to generate descriptions for individual classes in a single iteration. As a result, the generated descriptions tend to be overly general, with multiple categories sharing similar phrases, ~\emph{e.g.},~\emph{“various colors"} in bird classification. This generality hinders the ability to effectively discriminate between similar categories.

In light of these limitations, a key question arises: \textit{How can we design an automated pipeline that empowers LLMs to discover globally optimal class descriptions, thereby improving the overall visual classification performance?} In this paper, we introduce a novel approach, named Iterative Optimization with Visual Feedback, which demonstrates how an LLM agent can collaborate with VLMs, employing the feedback of visual classification to progressively refine class descriptions (Figure~\ref{fig1}). Our method formulates this task as a combinatorial optimization problem - identifying the combination of class descriptions for each category that maximizes VLM image classification performance. Given the problem's infinitely complex search space, we develop an LLM agent integrated with a Genetic Algorithm, where descriptions are evolved toward better solutions. Within each iterative cycle, the agent first conducts mutation based on the last round's descriptions and then performs crossover among various candidates to produce optimized concepts. This dual process of mutation and crossover allows the agent to explore the solution space both locally and globally, searching for the most effective visual concepts. We further introduce the concept of visual feedback to reduce variance across different results and computational resources, using image classification metrics from CLIP. Visual feedback can serve as both reward and memory for our agent, steering the LLM towards rational optimization and mitigating random-walk behavior during the process.

Extensive experiments conducted across nine image classification benchmark datasets reveal that our approach significantly outperforms current LLM-based methods as well as vanilla CLIP. We demonstrate that our LLM agent is able to iteratively discover highly descriptive visual descriptions that are conducive to image classification. We highlight another key insight that the final optimized class descriptions serve as robust visual knowledge with strong interpretability and transferability, which can consistently enhance model performance across different backbones and augment the efficacy of fine-tuning-based methods

In summary, our contributions are: \textbf{1)} We identify the limitations in existing methods and propose a novel paradigm for LLM-augmented visual classification by optimizing class descriptions with visual feedback from VLMs. \textbf{2)} To find the optimized descriptions, we design a genetic algorithm-inspired agent for iterative refinement without training. \textbf{3)} We conduct extensive experiments across 9 image classification datasets to validate the effectiveness of our optimized descriptions with additional explainability and transferability. The code of our methods and baselines are provided in the Supplementary Material.

\section{Related Work}
\label{sec:related}

\begin{figure*}[t]
   \centering
   \includegraphics[width=\textwidth]{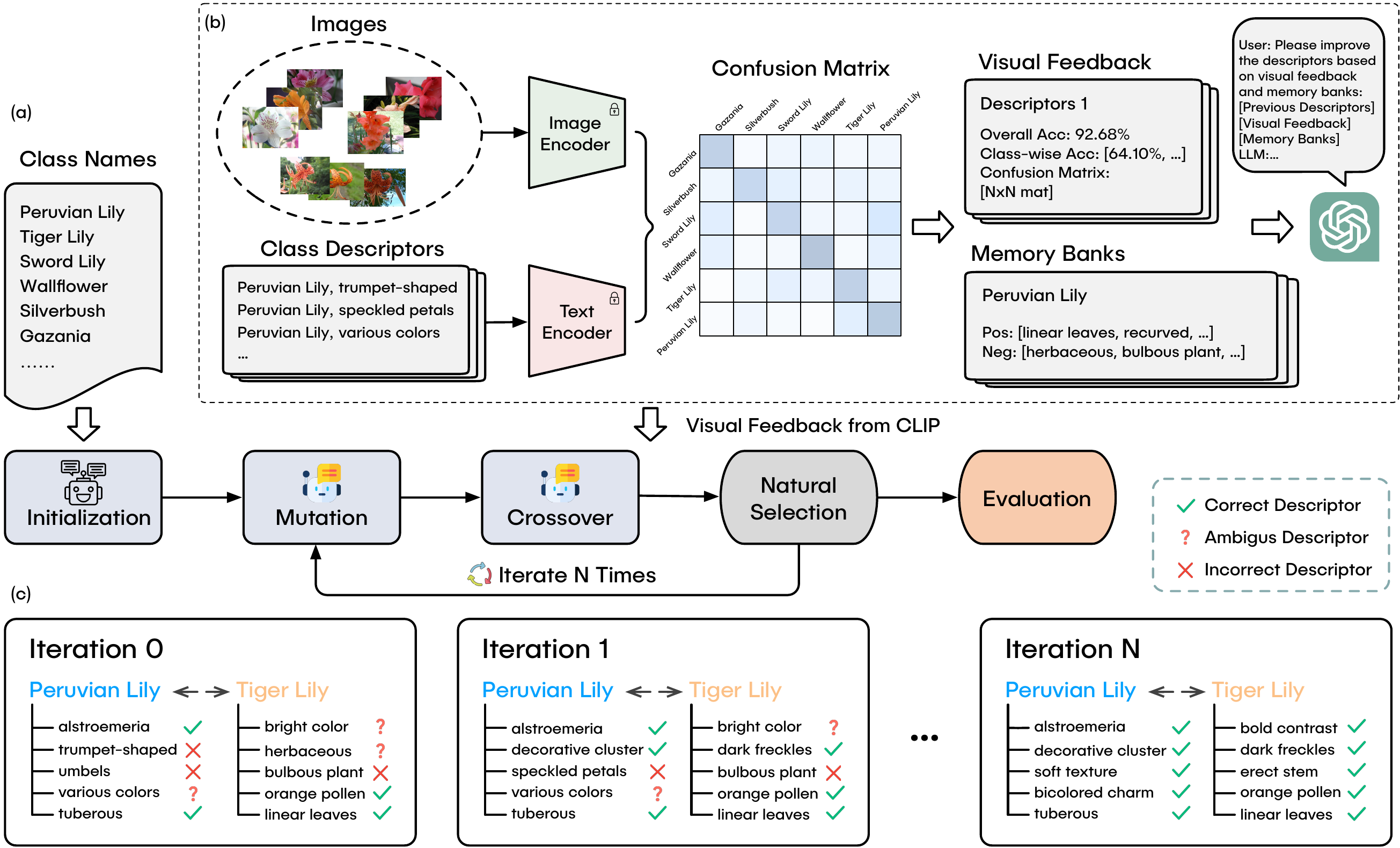} 
   \vspace{-4mm}
   \caption{\textbf{Illustration of iterative optimization with visual feedback.} (a) Given raw class names as input, we first prompt the LLM to generate an initialization of class descriptors. These descriptors undergo an iterative optimization comprising three stages: mutation, where diverse new candidates are generated; crossover, involving mixing and matching across different candidates to produce better candidates; and natural selection, selecting the most suitable candidate based on a fitness function. (b) In each iteration, we compute visual metrics including classification accuracy and confusion matrix for current class descriptors. We further use these metrics to construct visual feedback, update memory banks, and pick the best candidate in natural selection. (c) Through this iterative optimization, the LLM progressively identifies the most effective class descriptors, thereby enhancing the differentiation between ambiguous classes.}\vspace{-2mm}
   \label{mainfig}
\end{figure*}

\noindent\textbf{Large Language Models.}
Large Language Models (LLMs) have exhibited remarkable proficiency and sophisticated reasoning skills, significantly influencing various domains within artificial intelligence~\cite{bert,t5,gpt3,gpt4,llama,flan,palm}. These models have been proven capable of solving complex tasks, once thought to be solely within human capability, such as mathematical reasoning~\cite{cot,mathprompter}, drug discovery~\cite{drugchat,drugedit}, and decision makeing~\cite{yu2023language,eureka}. Their success in these areas underscores the planning and reasoning capabilities of LLMs. Furthermore, LLMs have shown immense potential in the multimodal domain~\cite{Kosmos1,minigpt4,llava,graphtext,nextgpt}. Most researchers align well-trained encoders from various modalities with LLMs through instruction tuning, equipping these models to interpret multimodal inputs. In contrast, our approach leverages a gradient-free method to integrate visual knowledge into LLMs without any need for fine-tuning. Nevertheless, the inherent limitations of LLMs in comprehending alternate modalities exacerbate the phenomenon of “hallucination" in multimodal contexts~\cite{woodpecker,aligning,holistic}, leading to inaccurate or even erroneous outputs. Thus, addressing this issue of hallucinations in LLMs represents a pivotal challenge.

\noindent\textbf{Prompt Engineering.}
Originating in NLP, prompt engineering significantly impacts the performance of VLMs in downstream tasks, leading many extensive research into identifying the optimal prompt. Prompt tuning~\cite{vpt,huang2022unsupervised,tip,coop,cocoop}, a method of parameter-efficient fine-tuning, involves introducing learnable parameters before the input text or image, which are then optimized through gradient updates. For instance, CoOp~\cite{coop} improves class descriptions by incorporating a set of parameters to represent dataset context, optimizing prediction accuracy via cross-entropy loss minimization. While prompt tuning notably increases accuracy, it necessitates additional training. Our method achieves comparable results without any training and serves as a complementary approach to prompt tuning, offering further precision improvements when applied subsequently.

\noindent\textbf{Using LLMs for Prompt Engineering.}
Recent advancements have seen the emergence of methods that employ LLMs to generate semantically richer descriptions for improving class prompts~\cite{DCLIP,WaffleCLIP,CuPL,CHiLS,yan2023learning}.~\citet{DCLIP} initially demonstrated that ensembling class-dependent descriptions generated by LLMs can improve classification accuracy. Building on this, WaffleCLIP~\cite{WaffleCLIP} incorporated high-level concepts related to the dataset to mitigate ambiguities in class names. CuPL~\cite{CuPL} utilizes a series of hand-crafted prompt templates to enable LLMs to produce diverse descriptions for each class. However, we observed that descriptions generated by these LLM-based methods often suffer from inaccuracies and ambiguities. Consequently, we propose a method involving iterative optimization to continuously refine descriptions, integrating visual feedback within the optimization process to enable LLMs to maximize differentiation between distinct categories.

\section{Method}
\label{sec:method}

\subsection{Classification with Descriptor Ensembling}
\label{sec:preliminary}
CLIP~\cite{CLIP} consists of an image encoder and a text encoder, which has been trained on 400M image-text pairs to learn a joint embedding space. Given a query image $x$ and a predefined set of classes $\mathcal{C} = \{c_1, c_2, c_3, ..., c_n\}$ in natural language, CLIP performs zero-shot image classification by first encoding both image and class names into the shared embedding space, then computing the cosine similarity between the image and each class, finally selecting the one with highest similarity as the predicted class,
\begin{equation}
    \tilde{c} = \argmax_{c\in \mathcal{C}} \cos(\phi_I(x), \phi_T(f(c))),
\end{equation}
where $\phi_I,\phi_T$ are the image encoder and text encoder, and $f(c)$ is the prompt template like \texttt{"A photo of a \{\textit{c}\}"}. 

Prior work~\cite{DCLIP} proposed a simple yet effective method to augment the class names $\mathcal{C}$ using LLMs. They prompt LLMs to generate a set of descriptors for each category, \emph{e.g.}, \textit{Hen: two legs; red, brown, or white feathers; a small body}. With these descriptive texts, they improve image classification accuracy by computing a comprehensive similarity score for each category: 
\begin{equation}
    \tilde{c} = \argmax_{c\in \mathcal{C}} \frac{1}{|\mathcal{D}(c)|} \sum_{d\in \mathcal{D}(c)} \cos(\phi_I(x), \phi_T(d)),
\end{equation}
where $\mathcal{D}(c)$ indicates the descriptors for class $c$, and we use $\mathcal{D}$ for short. By averaging the scores of all class descriptors, which achieves prompt ensembling, we argue that it reduces the noise of class name embedding and leads to more robust visual classification.

\subsection{Iterative Optimization with Visual Feedback}
\label{sec:main_method}
A key drawback of existing LLM-based methods is that they generate class descriptors in a single run, where LLM is frozen and not updated. By contrast, the ways humans recognize new objects always involve a dynamic learning process,~\emph{i.e.}, we gradually update our knowledge base of objects via interaction with the environment, remembering useful features and forgetting useless features. Inspired by that, we identify two fundamental points: interaction with the environment and iterative optimization, which are missing in existing methods. We formulate the problem of finding the optimal class descriptors as a combinatorial optimization problem. Further, we propose a novel method to dynamically optimize the set of class descriptors in an iterative manner integrated with visual feedback from CLIP.

Taking advantage of extended world knowledge and remarkable reasoning skills showcased by LLMs~\cite{gpt4,cot}, we empower LLMs to act as prompt-optimization agents to search for the best combination of class descriptors. Considering the complex solution space of possible combinations of class descriptors, we introduce an evolutionary process to search for optimal class descriptors. Classic genetic algorithms have been proven superior performance in solving complex optimization problems. The evolution usually starts with a population of randomly generated samples. In each iteration, it generates the next generation via two predefined operators,~\emph{mutation} and~\emph{crossover}, then selects the most promising offspring based on some fitness function that evaluates the quality of an individual. We illustrate the detailed process of each iteration in Algorithm~\ref{alg:main}, which will be explained in the following paragraphs.

\noindent\textbf{Initialization.}
Since the size of class labels can be large, especially for datasets like Imagenet~\cite{imagenet}, we first design a splitting strategy to group all classes into clusters based on the similarity of their names. We extract the text embedding of class names and then employ the K-means algorithm to cluster them into groups, where each group represents semantically similar classes. To make sure that LLM always focuses on confusing classes, the clustering step is dynamically conducted not only at initialization but also at the start of each iteration, where we compute the average embedding of all descriptors for each class. 

After clustering, we condition the LLM to generate initial class descriptors $\mathcal{D}_0$. Specifically, we instruct the LLM to generate $n_0$ descriptors for each class, providing task description, output formatting, and some design tips. We provide more implementation details about our prompts in the Supplementary Materials. The initialization step is similar to previous methods~\cite{DCLIP,WaffleCLIP,CuPL}. LLM can generate plausible descriptors usually related to colors, shapes, textures, etc. On the first try, however, LLM sometimes generates features that are ambiguous and irrelevant to visual classification due to the high degree of diversity and lack of specific pre-training data for visual understanding. Therefore, it is necessary to introduce both feedback from CLIP and iterative optimization to mitigate these issues.

\noindent\textbf{Visual Feedback.}
The key idea of visual feedback is grounding LLMs with visual knowledge in VLMs to better distinguish between ambiguous classes during the optimization process. Instead of directly updating the model parameters via instruction tuning in recent multimodal LLMs, we propose a gradient-free method to inject visual knowledge into LLMs. Specifically, given a current set of class descriptors $\mathcal{D}$, we construct visual feedback $V(\mathcal{D})$ through task-related evaluation metrics for CLIP, \emph{e.g.}, top-1 overall accuracy, class-wise accuracy, and confusion matrix in image classification. These metrics offer a holistic evaluation of model performance given current class descriptors. 

Apart from conventional metrics, we propose an improved version of the confusion matrix to more effectively capture intricate relationships within classes. We define a confusing threshold $\lambda$ and categorize each prediction as a positive sample based on its cosine similarity score compared to $\lambda$ times the cosine similarity score of the ground-truth label. In this approach, the improved confusion matrix, denoted as $\tilde{M}$, is computed by aggregating positive sample indicators. Another challenge is we find that LLMs sometimes struggle to interpret the raw confusion matrix with shape $|\mathcal{D}|*|\mathcal{D}|$, particularly as the number of classes $|\mathcal{D}|$ increases. Hence, we refine this process by extracting the top-$m$ classes from each row of $\tilde{M}$, representing the most confusing classes for CLIP. The visual feedback using the improved confusion matrix is formulated as follows:
\vspace{-4mm}

\begin{align}
\text{Pos}(x, d, d') &= \begin{cases}
1 & \text{if } \cos(x, d') > \lambda \cos(x, d) \\
0 & \text{otherwise}
\end{cases}, \\
\tilde{M}_{dd'} &= \sum_{x \in \mathcal{X}_d} \text{Pos}(x, d, d') \quad \text{for } d, d' \in \mathcal{D}, \\
V(\mathcal{D}) &= \bigcup_{d \in \mathcal{D}} \text{Top-}m\left(\tilde{M}_{d*}\right),
\end{align}
where $\mathcal{X}_d$ indicates the images of descriptor class $d$ to compute classification matrics in visual feedback, and we fix $\lambda=0.9, m=\frac{|\mathcal{D}|}{2}$ in all experiments. 

Though this metric-based visual feedback is simple to construct, it serves an important role in estimating the divergence between LLM-generated descriptors and optimal classification centroids in CLIP latent space. It has three major applications in our optimization process. First, we can convert the visual feedback $V(\mathcal{D})$ into natural language. During mutation and crossover, the textual version of $V(\mathcal{D})$ can be integrated into prompts to help LLM distinguish the target category from confusing classes. Second, we adopt $V(\mathcal{D})$ as the fitness function to evaluate sample quality and perform natural selection. Finally, we introduce the idea of memory banks $\mathcal{M}$ consisting of positive and negative history class descriptors. We dynamically update the memory banks based on $V(\mathcal{D})$ at the end of each iteration.

\begin{algorithm}[t]
    \caption{Interative Optimization with Visual Feedback}\label{alg:main}
    \begin{algorithmic}[1]
    \REQUIRE class labels $\mathcal{C}$, LLM \texttt{LLM}, prompt \texttt{prompt}, visual feedback $V$
    \item[\textbf{Hyperparameters:}] iterations $N$, mutation sample size $K$
    
        \vspace{3pt}
        \STATE Set $\mathcal{C} \gets \text{K-means}(\mathcal{C})$
        \vspace{3pt}
        \FOR{$i=1$ to $|\mathcal{C}|$}
            \STATE $\mathcal{D}_{0,i} = \texttt{LLM}(c_i, \texttt{prompt})$
        \ENDFOR
        \vspace{3pt}
        \STATE Set $\mathcal{M} \gets \emptyset$, $\Tilde{\mathcal{D}} \gets \mathcal{D}_0$
        \vspace{3pt}
        \FOR{$i=1$ to $N$}
            \STATE Set $\mathcal{D}_{i-1} \gets \text{K-means}(\mathcal{D}_{i-1})$
            \FOR{$j=1$ to $|\mathcal{D}_{i-1}|$}
                \STATE $\mathcal{D}_{i,j}^0, ..., \mathcal{D}_{i,j}^{K} = \texttt{LLM}(\texttt{prompt}, \mathcal{D}_{i-1,j}, \mathcal{M}, V)$
                \STATE $\mathcal{D}_{i,j}^{K+1} = \texttt{LLM}(\texttt{prompt}, \mathcal{D}_{i-1,j}, \mathcal{D}_{i,j}^0, ..., \mathcal{D}_{i,j}^{K}, V)$
                \STATE $\mathcal{D}_{i,j} = \argmax_{\mathcal{D} \in \{\mathcal{D}_{i,j}^0, ..., \mathcal{D}_{i,j}^{K+1}\}} V(\mathcal{D})$
            \ENDFOR
            \STATE $\mathcal{M} = \texttt{Update}(\mathcal{M}, \mathcal{D}_{i-1}, \mathcal{D}_{i}, V)$
            \STATE $\Tilde{\mathcal{D}} = \argmax_{\mathcal{D} \in \{\mathcal{D}_i\, \Tilde{\mathcal{D}}\}} V(\mathcal{D})$
        \ENDFOR
        \vspace{3pt}
    \OUTPUT $\Tilde{\mathcal{D}}$
\end{algorithmic}
\end{algorithm}

\noindent\textbf{Iterative Optimization.}
We first define the mutation and crossover operators to generate the next generation of class descriptors based on the previous version. For the $i$-th iteration, providing the previous set of class descriptors $\mathcal{D}_{i-1}$, $V(\mathcal{D}_{i-1})$, and memory banks $M$, we query LLM to pick the top $n_i$ most useless descriptors in the current set and replace them with $n_i$ new descriptors to emphasize its distinct visual features, representing the mutation operation. We generate $K$ independent candidates $\{\mathcal{D}_i^0, \mathcal{D}_i^1, ..., \mathcal{D}_i^{K}\}$ from LLM in each iteration to ensure sufficient genetic diversity for optimization. As for the crossover operation, we provide $K$ generated candidates in the mutation operation and their corresponding visual feedback as inputs then prompt LLM to perform mix and match between different samples, then output a new sample $\mathcal{D}_i^{K+1}$. The key idea for crossover is to ensemble different useful descriptors of different samples and produce an offspring with overall better performance. 

At the end of each iteration, we select the best performance candidate as $\mathcal{D}_i$ to update the current descriptor set among the population of generated class descriptors denoted as $\{\mathcal{D}_i^0, \mathcal{D}_i^1, ..., \mathcal{D}_i^{K+1}\}$. We adopt the overall accuracy in our visual feedback as the fitness score for natural selection. This fitness-based process ensures our method always chooses the best candidate as the starting point of the next iteration and produces class descriptors that better discriminate different categories, thus gradually moving towards the global optimal. To figure out the impact of descriptors and update the memory banks $\mathcal{M}$, we compare the difference of $\mathcal{D}_i$ and $\mathcal{D}_{i-1}$ in detail, resulting in three groups of descriptors,~\emph{i.e.}, unchanged, deleted, and added descriptors. Then we compute the visual feedback for these descriptors and compare their overall accuracy. If the accuracy of $\mathcal{D}_i$ is greater than that of unchanged descriptors, it indicates the added descriptors are beneficial for CLIP and we add them to the positive memory bank, otherwise, we add them to the negative one. Similarly, if the accuracy of $\mathcal{D}_{i-1}$ is greater than that of unchanged descriptors, it indicates the deleted descriptors are beneficial for CLIP and we add them to the negative memory bank, otherwise, we add them to the positive one.

\begin{table*}[t]
\centering
\caption{\textbf{Quality comparison of generated descriptions with other LLM-based methods and vanilla CLIP.} All data in the table represent top-1 accuracy (\%) on the test set, where bold figures indicate the highest accuracy consistently achieved by our method, while the underlined figures indicate the second-highest accuracy among the compared methods. $\Delta$ CLIP denote the absolute improvements of our method over vanilla CLIP.}
\resizebox{\textwidth}{!}{
\begin{tabular}{l|ccccccccc|c}
\toprule[1.5pt]
\multirow{2}{*}{\textbf{Method}} & \multicolumn{9}{c|}{\textbf{Dataset}} & \multirow{2}{*}{\textbf{Average}} \\

& \multicolumn{1}{c}{ImageNet} & \multicolumn{1}{c}{EuroSAT} & \multicolumn{1}{c}{UCF101}  & \multicolumn{1}{c}{SUN} & \multicolumn{1}{c}{Caltech} & \multicolumn{1}{c}{DTD}  & \multicolumn{1}{c}{CIFAR-10} & \multicolumn{1}{c}{Flowers102} & \multicolumn{1}{c|}{CUB} &  \\
\midrule
CLIP & 61.80 & 36.83 & 61.01 & 61.51 & 91.24 & 42.73 & 84.28 & 63.59 & 52.26 & 61.69\\
DCLIP & 63.00 & \underline{49.70} & 61.46 & 62.51 & 91.68 & 43.38 & 85.23 & \underline{67.21} & 53.25 & 64.16\\
WaffleCLIP & 62.83 & 49.69 & 60.88 & 63.65 & 89.29 & 43.97 & \underline{85.61} & 66.58 & \underline{53.47} & 64.00\\
CuPL & \underline{64.02} & 48.05 & \underline{63.26} & \underline{64.74} & \underline{91.72} & \underline{46.04} & 85.29 & 65.25 & 53.21 & \underline{64.62}\\
\textbf{Ours} & \textbf{64.53} & \textbf{56.28} & \textbf{67.01} & \textbf{66.22} & \textbf{92.70} & \textbf{51.42} & \textbf{86.33} & \textbf{72.19} & \textbf{56.13} & \textbf{68.09}\\
\midrule
$\Delta$ CLIP & \textcolor{teal}{$+2.73$} & \textcolor{teal}{$+19.45$} & \textcolor{teal}{$+6.00$} & \textcolor{teal}{$+4.71$} & \textcolor{teal}{$+1.46$} & \textcolor{teal}{$+8.69$} & \textcolor{teal}{$+2.05$} & \textcolor{teal}{$+8.60$} & \textcolor{teal}{$+3.87$} & \textcolor{teal}{$+6.40$} \\
\bottomrule[1.5pt]
\end{tabular}
}
\vskip -0.1in
\label{tab:sota}
\end{table*}

\subsection{Comparison with Other Methods}
\label{sec:comparison}
Our method distinguishes itself from existing LLM-based techniques~\cite{DCLIP,WaffleCLIP,CuPL} in its reliance on labeled image data for optimization, deviating from the conventional zero-shot image classification paradigm. However, the ultimate goal of our method is to obtain an optimal set of class descriptions for the target dataset. Considering the prevalent issues of ambiguity and inaccuracy in category descriptions generated by LLMs in a zero-shot manner, the integration of VLMs into the optimization process is essential. Our experiments demonstrate that the class descriptions derived through our method more effectively enhance CLIP in image classification and yield comparable results even in a low-shot setting.

Furthermore, our method also significantly diverges from fine-tuning-based approaches~\cite{coop,cocoop}. Our approach is training-free, thereby avoiding potential overfitting issues, which is orthogonal to these works. Our empirical results underscore the robustness and transferability of the optimized class descriptions, demonstrating their efficacy across various model backbones. This adaptability suggests promising applications, including integrating these descriptions into fine-tuning processes to further elevate their performance and generalizability, a claim we substantiate in our experimental section.

\section{Experiments}
\label{sec:exp}

\subsection{Experimental Setup}

\noindent\textbf{Implementation Details.}
In our method, there are four main hyperparameters, including the number of iterations $N$, the number of descriptors at initialization $n_0$, the number of descriptors to change in mutation and crossover $n_i$, and the number of mutated candidates in each iteration $K$. 
We set $N=10, n_0=30, n_i=15, K=4$ for all datasets. As for the number of groups in K-means clustering, we set it to $\texttt{round}(\frac{|\mathcal{C}|}{10})$ to make sure there are roughly $10$ classes for each group. Unless specified, we adopt GPT-4~\cite{gpt4} (with temperature fixed at 1.0) to construct our LLM agent and CLIP ViT-B/32 backbone~\cite{CLIP} to extract the image and text embeddings.

\noindent\textbf{Datasets.} 
Our experiments leverages the dataset partitioning introduced by CoOp~\cite{coop} on 9 different image classification benchmarks, including: ImageNet~\cite{imagenet}, EuroSAT~\cite{eurosat}, UCF101~\cite{ucf101}, Scene UNderstanding (SUN)~\cite{sun}, Caltech~\cite{caltech}, Describable Textures Dataset (DTD)~\cite{dtd}, CIFAR-10~\cite{cifar10}, Flowers102~\cite{flowers}, and CUB~\cite{CUB}.

\noindent\textbf{Compared Methods.} We compare our work with vanilla CLIP and three state-of-the-art methods using LLM to augment class descriptions. \textbf{CLIP}~\cite{CLIP} sets a simple template as \texttt{"A photo of a \{class name\}"} as input prompt. \textbf{DCLIP}~\cite{DCLIP} usess LLM-generated class descriptors with a few in-context examples to improve CLIP classification. They build the prompt in the format of \texttt{"\{classname\}, which (is/has/etc) \{descriptor\}"}. \textbf{WaffleCLIP}~\cite{WaffleCLIP} further improves DCLIP by introducing high-level concepts at the beginning of the prompt and replacing class descriptors with random characters. The extended prompt is \texttt{"A photo of a \{concept\}: a \{classname\}, which (is/has/etc) \{random\_sequence\}"}. \textbf{CuPL}~\cite{CuPL} designs hand-crafted prompts for LLMs to generate diverse descriptive sentences.

As clarified in Section~\ref{sec:comparison}, our method utilizes label information of training samples to guide the optimization process. Therefore, we are not comparing the above methods in the conventional zero-shot setting. Instead, we aim to directly compare the effectiveness of the generated descriptions via the evaluation of image classification. 

\subsection{Main Result}
As shown in Table~\ref{tab:sota}, the quality of the descriptions generated by our method surpasses that of existing LLM-based methods, as validated across a wide range of datasets, and is significantly higher than the category labels of vanilla CLIP. Notably, our method achieves an average increase of over $6\%$ compared to CLIP. Our method excels in relatively abstract, visually challenging datasets, such as the texture classification dataset, DTD, and the satellite image classification dataset, EuroSAT, with $8.69\%$ and $19.45\%$ absolute improvements, respectively. For datasets with finer category granularity, like Flowers102, CUB, UCF101, and SUN, our method also outperforms CLIP by over $3.5\%$, with a remarkable $8.6\%$ increase on the Flowers102 dataset. However, other LLM-based methods' best performances on these datasets are limited. This highlights our method's superiority in finding better descriptions of various datasets, where multi-round iterations and visual feedback are essential in distinguishing closely related and confusing categories. This is not achievable in single-category and single-turn optimization methods like DCLIP, WaffleCLIP, and CuPL. Without visual knowledge in pre-trained VLMs, these methods generate ambiguous and inaccurate descriptions in a single run, providing no discriminative information, ~\emph{e.g.}, \textit{“black bill"} for most classes of birds in CUB. On datasets like CIFAR-10, Caltech, and ImageNet, the margins between our approach and other LLM-based baselines are relatively smaller. We attribute this to the fact that these datasets already include distinct classes, which are easier for LLMs to recognize and understand through class names.

\begin{figure}[t]
   \centering
   \includegraphics[width=\linewidth]{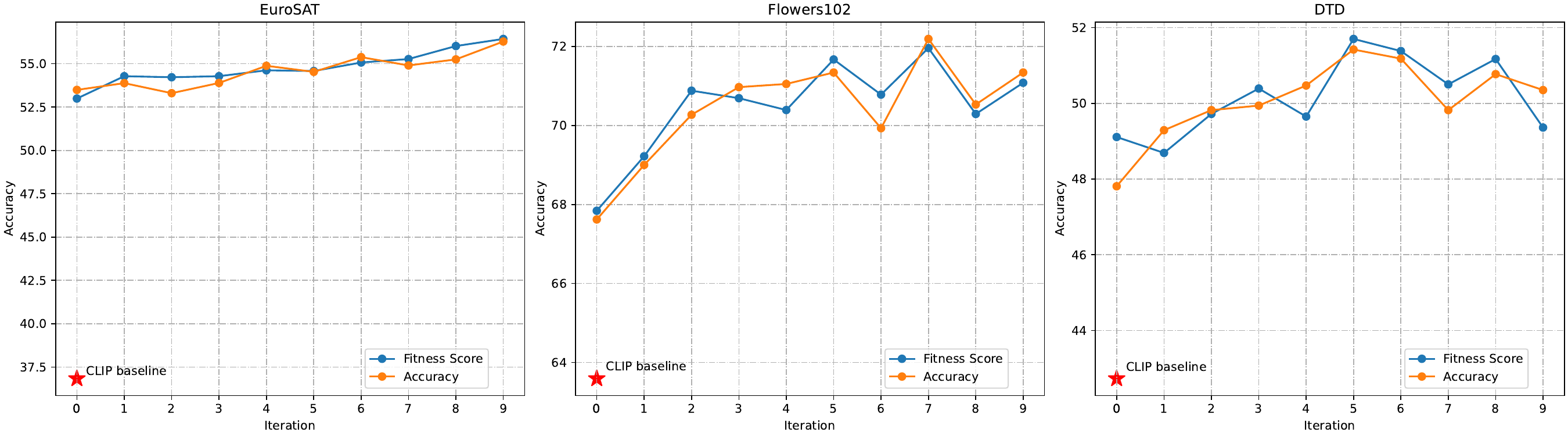}
   \vspace{-6mm}
   \caption{\textbf{Ablation on iterative optimizations.} X-axis: iteration rounds, Y-axis: Accuracy (\%). Red stars represent the accuracy of vanilla CLIP.}
   \label{figabl1}
   \vspace{-4mm}
\end{figure}

\subsection{Ablation Study}
\label{sec:ablation}
\noindent\textbf{Iterative Optimization.} First, we discuss the impact of iterative optimization. Figure~\ref{figabl1} shows that as the number of iterations increases, both fitness score and test accuracy exhibit an initial growth followed by stable oscillations. After iterations of optimization, the final results significantly surpass the performance of CLIP baseline and single-turn methods at iteration $0$. Experimental results indicate that adopting our genetic algorithm-inspired optimization effectively moves the class descriptors toward better solutions, advancing the overall classification accuracy. Furthermore, the trends of the two lines in each graph are closely aligned, showcasing that there are no “overfitting" issues during optimization. This emphasizes the potential of our final descriptors as a universal textual feature across different model backbones, which is proven in the following text.

\begin{table}[t]
\centering
\caption{\textbf{Ablation on visual feedback.}}
\resizebox{0.8\linewidth}{!}{%
\begin{tabular}{lccc}
\toprule[1.5pt]
   \multicolumn{1}{l}{Method} &\multicolumn{1}{c}{Caltech} &\multicolumn{1}{c}{EuroSAT} &\multicolumn{1}{c}{Flowers102}   \\
\midrule
Ours(iCM) & \textbf{92.01} & \textbf{52.26} & \textbf{71.34}\\
Ours(CM) & 91.98 & 48.14 & 70.47\\
w/o Memory & 90.89 & 50.59 & 67.97\\ 
w/o Feedback & 91.11 & 47.16 & 68.32\\
\bottomrule[1.5pt]
\end{tabular}%
}
\vskip -0.15in
\label{tab:ablation}
\end{table}

\begin{table*}[t]
\centering
\caption{\textbf{Transferring optimized descriptors to different model architectures.}}
\resizebox{0.95\textwidth}{!}{%
\begin{tabular}{lllllllllllll}
\toprule[1.5pt]
   & \multicolumn{3}{c}{ImageNet} & \multicolumn{3}{c}{EuroSAT} & \multicolumn{3}{c}{Caltech} & \multicolumn{3}{c}{Flowers102}   \\
 
CLIP Architecture & \multicolumn{1}{c}{Ours} & \multicolumn{1}{c}{CLIP} & \multicolumn{1}{c}{$\Delta$} & \multicolumn{1}{c}{Ours} & \multicolumn{1}{c}{CLIP} & \multicolumn{1}{c}{$\Delta$} & \multicolumn{1}{c}{Ours} & \multicolumn{1}{c}{CLIP} & \multicolumn{1}{c}{$\Delta$} & \multicolumn{1}{c}{Ours} & \multicolumn{1}{c}{CLIP} & \multicolumn{1}{c}{$\Delta$} \\ 
\cmidrule(lr){1-1}\cmidrule(lr){2-4}\cmidrule(lr){5-7}\cmidrule(lr){8-10}\cmidrule(lr){11-13}
ViT-B/32 & \textbf{64.53} & 61.80 & \textcolor{teal} {+2.73}  & \textbf{56.28} & 36.83 & \textcolor{teal} {+19.45}  & \textbf{92.70} & 91.24 & \textcolor{teal} {+1.46}  & \textbf{72.19} & 63.59 & \textcolor{teal} {+8.60}  \\
\midrule
RN101 & \textbf{63.36} & 60.65 & \textcolor{teal} {+2.71}  & \textbf{36.30} & 32.02 & \textcolor{teal} {+4.28}  & \textbf{91.81} & 89.49 & \textcolor{teal} {+2.32}  & \textbf{70.85} & 61.39 & \textcolor{teal} {+9.46} \\

ViT-B/16 & \textbf{69.51} & 66.63 & \textcolor{teal} {+2.88}  & \textbf{52.09}  & 42.96 & \textcolor{teal} {+9.13}  & \textbf{94.48} & 92.54 & \textcolor{teal} {+1.94}  & \textbf{75.48} & 66.46 & \textcolor{teal} {+9.02}  \\
ViT-L/14 & \textbf{76.11} & 72.85 & \textcolor{teal} {+3.26}  & \textbf{67.40} & 52.86 & \textcolor{teal} {+14.54}  & \textbf{96.80}  & 94.04 & \textcolor{teal} {+2.76} & \textbf{81.73} & 75.96 & \textcolor{teal} {+5.77} \\ 
\bottomrule[1.5pt]
\end{tabular}%
}
\vspace{-2mm}
\label{tab:transfer}
\end{table*}

\begin{figure*}[t]
   \centering
   \includegraphics[width=\linewidth]{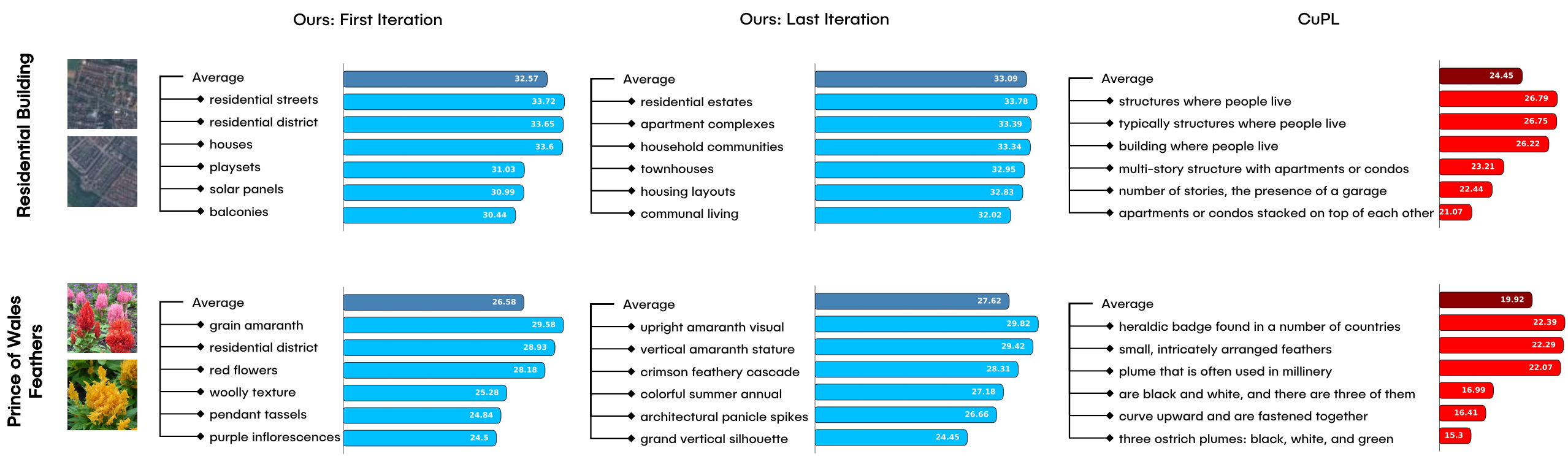}\vspace{-2mm}
   \caption{\textbf{Examples of interpretability.} We select two categories from EuroSAT and Flowers102 and list the top-$3$ and last-$3$ descriptions for each category, ranked by their similarity scores averaging across class test samples.}\vspace{-2mm}
   \label{fig:visual}
\end{figure*}

\noindent\textbf{Visual Feedback.}
Next, we study the importance of each component of visual feedback in our method. As illustrated in Table~\ref{tab:ablation}, we design four baselines for ablations: 1) Ours with improved confusion matrix (iCM). This is our standard setting, where we adopt both the improved confusion matrix as visual feedback and memory banks to enhance descriptor optimization described in Section~\ref{sec:main_method}. 2) Ours with confusion matrix (CM). This baseline simply replaces the improved confusion matrix with the conventional confusion matrix. 3) Without Memory. We remove the memory bank recording iteration history, but the improved confusion matrix is kept. 4) Without Feedback. In this setting, we remove all components related to visual feedback, including both the confusion matrix and memory banks.

As shown in Table~\ref{tab:ablation}, the use of the improved confusion matrix demonstrates the best performance on three datasets. The performance with the standard confusion matrix is second best, highlighting the effectiveness of the memory bank and visual feedback components designed in our work. Notably, removing either the memory bank or the visual feedback components individually leads to a decrease in generation accuracy and instability. For detailed graphs, please refer to the appendix~\ref{sec:visual-feedback}, where our method shows the most stable improvements toward better descriptions.

\noindent\textbf{Few-shot Setting.} Our few-shot ablation experimental results are shown in the appendix~\ref{sec:few-shot}, demonstrating that high-quality class descriptions can be obtained using our method even with a very small proportion of samples.

\subsection{Transferability}
\noindent\textbf{Transfer to Different Backbones.}
To verify the generalization and transferability of our method, we tested the optimized descriptions obtained using CLIP ViT-B/32 on different backbones, including RN101, ViT-B/16, and ViT-L/14. Our results, presented in Table~\ref{tab:transfer}, demonstrate that our optimized class descriptions consistently surpass the baseline CLIP model in terms of accuracy. A notable observation is the significant performance enhancement on the Flowers102 dataset, with increases exceeding $9\%$ when transferred to RN101 and ViT-B/16. These results emphasize our method generates more generalizable natural language prompts, avoiding overfitting to a specific architecture.

\begin{table}[t]
\centering
\caption{\textbf{Fine-tuning results using generated descriptions.}}
\resizebox{\linewidth}{!}{%
\begin{tabular}{lcccccc}
\toprule[1.5pt]
Setting & Ours & ImageNet & EuroSAT & UCF & DTD & Flowers102 \\ 
\midrule[1pt]
\multirow{2}{*}{1 shot} & \ding{55} & 62.6 & 46.0 & 65.6 & 53.1 & 72.8 \\
& \checkmark & 64.8 & 56.9 & 67.9 & 56.0 & 74.2 \\
\midrule
\multirow{2}{*}{2 shot} & \ding{55} & 62.4 & 55.5 & 68.6 & 54.6 & 75.9 \\
& \checkmark & 64.3 & 62.7 & 70.2 & 57.9 & 77.1 \\
\midrule
\multirow{2}{*}{4 shot} & \ding{55} & 63.6 & 58.4 & 73.2 & 57.3 & 84.9 \\
& \checkmark & 65.1 & 61.5 & 73.7 & 57.4 & 82.6 \\
\midrule
\multirow{2}{*}{8 shot} & \ding{55} & 65.0 & 63.8 & 76.9 & 61.3 & 90.2 \\
& \checkmark & 65.8 & 68.7 & 77.3 & 61.9 & 90.1 \\
\midrule
\multirow{2}{*}{16 shot} & \ding{55} & 66.6 & 74.3 & 81.1 & 66.6 & 94.0 \\
& \checkmark & 67.0 & 75.4 & 81.2 & 67.6 & 93.9 \\
\bottomrule[1.5pt]
\end{tabular}
}\vskip -0.15in
\label{tab:ablation-ft}
\end{table}

\noindent\textbf{Transfer to Fine-tuning Setting.} We further demonstrate that our optimized descriptions can be transferred to fine-tuning-based approaches for CLIP. We take CLIP-A-Self~\cite{iccvw} as an example, which employs a self-attention-based adapter for feature selection and aggregation. As indicated in Table~\ref{tab:ablation-ft}, integrating descriptions generated by our method into the CLIP-A-Self framework results in notable performance enhancements across all datasets. Moreover, our method exhibits a clear advantage in low-shot learning scenarios.

\subsection{Interpretability and Analysis}
\label{sec:visualiation}
Figure~\ref{fig:visual} visualizes LLM-generated descriptions that contribute the most and least for classifying images of different classes. Specifically, we select our results at the first and last iteration, as well as results from CuPL~\cite{CuPL} for visualization, with class Residential Building in EuroSAT and Prince of Wales Feathers in Flowers102. To compare the effectiveness of descriptions, we compute the similarity score for each description using the average of all test images in that category. After sorting them in descending order, we visualize the top-$3$ and last-$3$ descriptions and their similarity scores in each setting.

Due to the high degree of diversity in CuPL, the quality of generated descriptions greatly varies, and it is prone to generate vague and overlapping descriptions,~\emph{e.g.}, \textit{“structures where people live"} and \textit{“building where people live"}. In contrast, our method generates more concise and relevant descriptions at initialization and continues to refine them during optimization, showing consistent growth in similarity scores of all descriptions. Our final descriptions reveal strong class-dependant semantic information which is beneficial for visual classification. In addition, for the confusing class name “Prince of Wales Feathers" in Flowers102, CuPL mistook it as the heraldic badge of the Prince of Wales instead of an annual herb, generating completely irrelevant descriptions. We further find this issue exists in all LLM-based methods. Instead, our method successfully generates correct descriptions since we provide the information of related classes and leverage feedback from CLIP for visual grounding, emphasizing the importance of visual feedback.

\vspace{-2mm}
\section{Conclusion}
\label{sec:conclusion}

In this work, we present a novel paradigm for image classification that leverages LLMs to iteratively refine class descriptors with visual feedback from VLMs to guide the optimization process. 
We validate the effectiveness of optimized descriptions by our method across 9 image classification datasets, showcasing superior performance with multiple benefits including interpretability and transferability.
\section{Broader Impact}
The integration of Large Language Models and Vision-Language Models for optimized image classification, as discussed in the paper, presents both ethical and social implications. Ethically, it enhances accuracy and explainability in critical applications like medical diagnostics, raising concerns about potential biases in training data affecting fairness. Socially, the increased interpretability of models may foster greater trust and wider adoption in various sectors. However, it also poses risks related to privacy and misuse in surveillance. Addressing these challenges necessitates future research focused on ensuring fairness, privacy, and transparency.

\bibliographystyle{icml2024}
\bibliography{reference}

\newpage
\appendix
\onecolumn
\setcounter{page}{1}
\setcounter{section}{0}
\renewcommand{\thesection}{\Alph{section}}
\section{Ablation on Few shot Study}
\label{sec:few-shot}
\begin{table}[h]
\centering
\caption{\textbf{Ablation on the number of shots used for calculating the confusion matrix.}}
\vskip 0.15in
\resizebox{0.5\linewidth}{!}{%
\begin{tabular}{lcccc}
\toprule[1.5pt]
   \multicolumn{1}{l}{Setting} &\multicolumn{1}{c}{UCF} &\multicolumn{1}{c}{Caltech} &\multicolumn{1}{c}{Flowers102}  &\multicolumn{1}{c}{CUB}  \\
\midrule
1 shot & 64.26 & 89.68 & 69.07 & 54.87\\
2 shot & 64.31 & 90.81 & 68.45 & 55.28\\
4 shot & 65.19 & 90.68 & 70.48 & 55.37\\ 
8 shot & 66.14 & 91.04 & 69.54 & 55.37\\
16 shot & 65.27 & 91.33 & 70.65 & 55.54\\
\bottomrule[1.5pt]
\end{tabular}%
}
\vspace{-1.5mm}
\label{tab:ablation-shot}
\end{table}

In this section, we present the experimental results of a few-shot study, as shown in the table. The term ‘‘shot'' refers to the number of labeled samples used per category to compute visual feedback. However, our few-shot setting is different from conventional ones since we do not require any training. Overall, the quality of the generated descriptions improves as the number of shots increases. Even with a very limited number of shots, it is possible to achieve satisfactory performance.

\section{Iterative Optimization Visualization}
\label{sec:result-example}
\begin{figure*}[h]
    \centering
    \includegraphics[width=\linewidth]{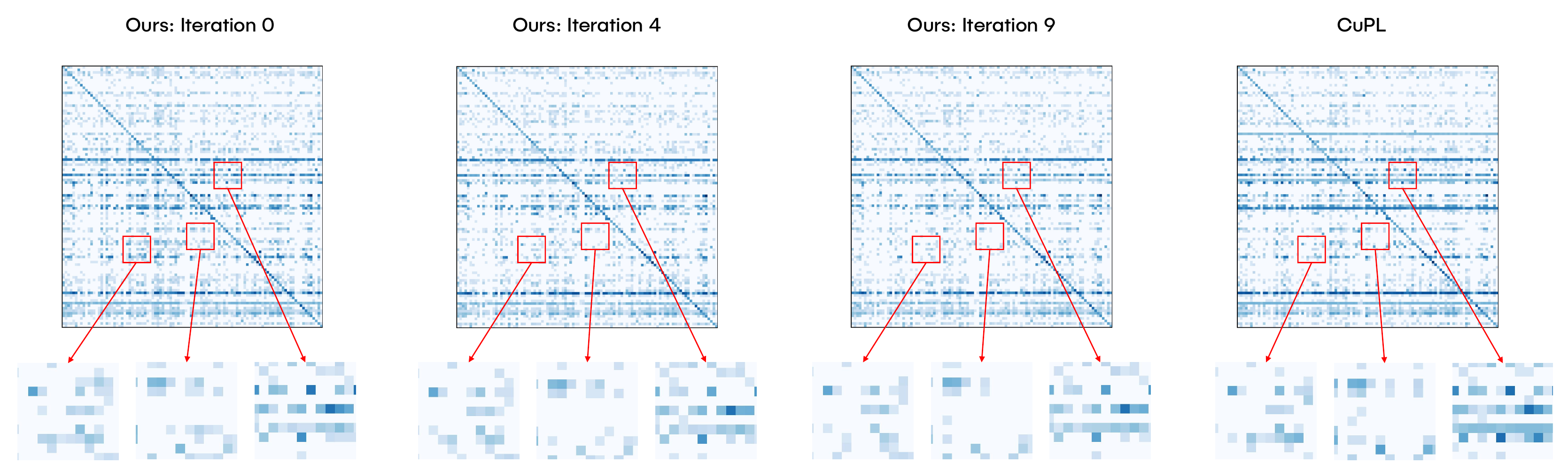}
    \caption{\textbf{Iterative Optimization Visualization.} From left to right, the sequence is as follows: the $0$-th round of our method, the $4$-th round of our method, the $9$-th round of our method, and the CuPL~\cite{CuPL} method. The darker the color in the heatmap, the higher the corresponding value in the confusion matrix.}
    \label{fig:supvis}
\end{figure*}
To further demonstrate the superiority of iterative optimization, in this section, we present the confusion matrix as heatmaps, offering a more intuitive visualization of the improvements brought about by the optimization process and a comparison with a previous state-of-the-art method. As shown in Figure~\ref{fig:supvis}, the three images on the left respectively represent the initial, mid-stage, and final-round confusion matrix heatmaps of our method, while the image on the far right is the heatmap of the confusion matrix generated using the CuPL~\cite{CuPL} method for description generation. In our approach, as seen in the three randomly selected areas, the color intensity in the non-diagonal regions lightens as the iterations increase. This indicates a reduction in the number of categories confused with the diagonal class over time, further underscoring the effectiveness of iterative optimization. Comparing our method with CuPL, we notice many rows in the heatmap are much darker, indicating a severe blurring of class distinction along the diagonals. This also reveals that the strategy of using LLMs to generate detailed categories in previous methods has limited impacts on optimizing classification effectiveness.

\section{Visual Feedback Experiment}
\label{sec:visual-feedback}
\begin{figure*}[h]
   \centering
   \includegraphics[width=\textwidth]{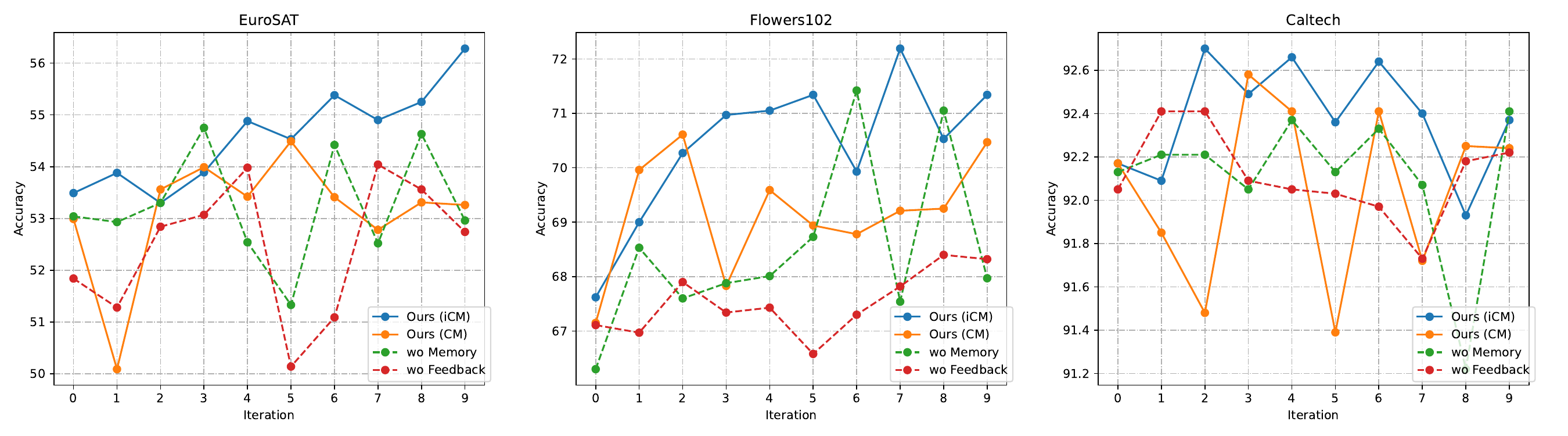}
   \vspace{-3mm}
   \caption{\textbf{Ablation on visual feedback.} X-axis: iteration rounds, where iteration $0$ indicates initialization, Y-axis: Accuracy (\%). “iCM" indicates improved confusion matrix, “CM" indicates confusion matrix, “wo Memory" indicates without memory banks, and “wo Feedback" indicates without all components related to visual feedback.}
   \label{figabl2}
\end{figure*}
In Figure~\ref{figabl2}, solid lines represent the complete setup, while dashed lines indicate scenarios where components of visual feedback are omitted. Comparing the two solid lines, the improved confusion matrix demonstrates overall superior performance to the standard confusion matrix. The conventional confusion matrix not only contains excessive redundant information, which may hamper the understanding of LLMs, but also discards certain critical information to discriminate related classes since it relies solely on the top-1 accuracy for construction. Our improved version effectively addresses these issues. Examining the dashed line for “without memory" alongside the two solid lines showcases that removing the memory bank leads to significant fluctuations in accuracy. Although occasionally reaching relatively high accuracy, such a model contains high variances but lacks robustness. Hence, the memory bank significantly enhances the stability and robustness during the optimization process. Regarding the setting without feedback, which is the removal of both the confusion matrix and memory banks, we notice a minimal increase or even drop in accuracy, reaffirming the crucial role of visual feedback.

\section{Full Prompts}
\label{sec:full-prompts}
In this section, we provide full prompts for our proposed method of iterative optimization with visual feedback. There are three components in this framework: initialization, mutation, and crossover. Each component has a system prompt and a user prompt.

\begin{lstlisting}[basicstyle=\fontfamily{\ttdefault}\scriptsize, breaklines=true,caption={\textbf{Initialization system prompt.}}]
You are a prompt engineer trying to optimize the text description of class labels for image classification. The CLIP 
model performs zero-shot image classification by computing the cosine similarities between input images and class l-
abels. You are given original class labels and some hints of confusion classes that are indistinguishable to CLIP f-
or each class. Your goal is to generate a list of visual concepts to improve the description of current class labels 
to maximize their distinctions for CLIP to better recognize. 
The output class names should be a Python list of strings in the following format:
```
["original_class_name1: concept1, concept2, concept3, ...", "original_class_name2: concept1, concept2, concept3, 
...", ...]
```

Some helpful tips for optimizing the class descriptions:
    1. You should generate {n_concepts_init} different high-level concept words for each class to emphasize the 
    distinct visual features of this class but not appear in other classes, split them with ",".
    
    2. DO NOT give me non-visual words!
    
    3. The class description must also be general enough to cover most training images of this class.
    
    4. Most importantly, always focus on the visual features of all classes, since there are only images input. Do 
    not produce text describing invisible features, e.g., voice, mental character, etc.
    
    5. Do not include any other class names in the description of one class, which means do not use text like "not 
    like a cat" in dogs or "distinct from birds"in plans. Only focus on the features of the class itself.
    
    6. The concept words for each class should be diverseand not too similar to each other.
\end{lstlisting}
\begin{lstlisting}[basicstyle=\fontfamily{\ttdefault}\scriptsize, breaklines=true,caption={\textbf{Initialization user prompt.}}]
The original class names are:
{current_class}

The confusion classes for CLIP are:
{visual_feedback}
Note, classes in the front are more prone to be confused than the ones in the back.

Write the optimized class names leveraging the hints of confusion classes list above. Directly generate your answer 
in the given format without any extra information.
\end{lstlisting}
\begin{lstlisting}[basicstyle=\fontfamily{\ttdefault}\scriptsize, breaklines=true,caption={\textbf{Mutation system prompt.}}]
You are a prompt engineer trying to optimize the text description of class labels for image classification. The CLIP 
model performs zero-shot image classification by computing the cosine similarities between input images and class l-
abels. You are given current class labels, some hints of confusion classes that are indistinguishable to CLIP and l-
ists of good and bad history words for each class. Your goal is to optimize the description of current class labels 
to maximize their distinctions for CLIP to better recognize images.
The output class names should be a Python list of strings in the following format:
```
["original_class_name1: concept1, concept2, concept3, ...", "original_class_name2: concept1, concept2, concept3, 
...", ...]
```

Some helpful tips for optimizing the class descriptions:
    1. Base on all the provided information, you should identify {n_concepts} bad high-level concept words in each 
    class and replace them with better ones to emphasize the distinct visual feature of this class but not appear in
    others', split them with ",".
    
    2. DO NOT give me non-visual words!
    
    3. The class description must also be general enough to cover most training images of this class.
    
    4. Most importantly, always focus on the visual features of all classes, since there are only images input. Do 
    not produce text describing invisible features, e.g., voice, mental character, etc.
    
    5. Do not include any other class names in the description of one class, which means do not use text like "not 
    like a cat" in dogs or "distinct from birds" in plans. Only focus on the features of the class itself.
    
    6. The concept words for each class should be diverse and not too similar to each other.
\end{lstlisting}
\begin{lstlisting}[basicstyle=\fontfamily{\ttdefault}\scriptsize, breaklines=true,caption={\textbf{Mutation user prompt.}}]
The current class names are:
{current_class}

The confusion classes for CLIP are:
{visual_feedback}
Note, classes in the front are more prone to be confused than the ones in the back.

History list class descriptions that will improve classification accuracy are:
{positive_list}

History list records for class descriptions that will decrease classification accuracy are:
{negative_list}

Write the optimized class names leveraging the hints of confusion classes and two lists above. Directly generate your answer in the given format without any extra information.
\end{lstlisting}
\begin{lstlisting}[basicstyle=\fontfamily{\ttdefault}\scriptsize, breaklines=true,caption={\textbf{Crossover system prompt.}}]
You are a prompt engineer trying to optimize the text description of class labels for image classification. The CLIP 
model performs zero-shot image classification by computing the cosine similarities between input images and class l-
abels. You are given {n_samples} versions of class labels with different text descriptions, and classification metr-
ics for each of them. Your task is to find the best combination of given text description for each class label based 
on the classification metrics to achieve the best overall classification performance.
The output class names should be a Python list of strings in the following format:
```
["original_class_name1: concept1, concept2, concept3, ...", "original_class_name2: concept1, concept2, concept3, 
...", ...]
```

Some helpful tips for optimize the class descriptions:
    1. You should only select from the existing concept words in class descriptions. You cannot create new words.
    
    2. For each class, you should combine various concepts in different versions of descriptions, and generate 
    the description that is most conducive to distinguishing this class.
    
    3. You should leverage the classification metrics for your selection. The overall accuracy indicates the global 
    performance, while the class-wise accuracy indicates the performance for each class.
    
    4. Keep the number of concepts in each class description unchanged, i.e., {n_concepts_init} concepts for each 
    class.
\end{lstlisting}
\begin{lstlisting}[basicstyle=\fontfamily{\ttdefault}\scriptsize, breaklines=true,caption={\textbf{Crossover user prompt.}}]
The {n_samples} versions of class descriptions and corresponding classification metrics are:{class_samples}

Output the optimized class descriptions according to the classification metrics above. Directly generate your answer 
in given format without any extra information.
\end{lstlisting}

\section{Result Example}
\label{sec:result-example}
In this section, we showcase some examples generated by our method. We select a subset of class labels from two datasets for demonstration. For the complete set of labels, please refer to the \texttt{.txt} file in our Supplementary Material.
\begin{lstlisting}[basicstyle=\fontfamily{\ttdefault}\scriptsize, breaklines=true,caption={\textbf{Examples of generated descriptions for Flowers102.}}]
{
    "tiger lily": [
        "erect perennial growth",
        "pollen-covered anthers",
        "intense orange hue",
        "lance-shaped leaf structure",
        "vibrant orange-red coloring",
        "summer garden fixture",
        "bold tiger-striped petals",
        "straight growth form",
        "midsummer peak",
        "lanceolate leaf shape"
    ],
    "giant white arum lily": [
        "robust flora",
        "monochromatic palette",
        "tropically adapted",
        "arrow-shaped leaves",
        "swamp inhabitant",
        "stately appearance",
        "massive spoon-like spathe",
        "lush marsh foliage",
        "imposing white florescence",
        "prominent central spadix",
        "large ovate bracts",
        "waxy texture",
        "South African native",
        "water-adjacent growth"
    ],
    "fire lily": [
        "south african endemic",
        "scorching colors",
        "wild habitat specialist",
        "volcanic color palette",
        "excessive pollen attractor",
        "specific African habitat",
        "nectar-abundant flower",
        "curved petal silhouettes"
    ],
    "orange dahlia": [
        "horticultural pride",
        "summer crescendo",
        "elegant growth habit",
        "fiery orange inflorescence",
        "lush verdant foliage",
        "decorative cutting flower",
        "opposite leaf arrangement",
        "compacted petal rows",
        "bold spherical blooms",
        "mid-summer grandeur",
        "fiery",
        "ornamental",
        "foliage"
    ],
    "pink-yellow dahlia": [
        "cut flower favorite",
        "standout horticultural beauty",
        "large petal-packed inflorescence",
        "bedding plant",
        "luminous garden feature",
        "autumn flowering",
        "ornamental garden treasure",
        "delicate pink-yellow gradient petals",
        "soft petal curvature",
        "opulent look",
        "gradient",
        "autumn",
        "opulent"
    ],
    "cautleya spicata": [
        "woodland understory preference",
        "shade-loving",
        "soft yellow tones",
        "Himalayan forest dweller",
        "orchid resemblance",
        "clusters of golden flowers",
        "semi-evergreen",
        "rhizomes rooting",
        "delicate mountainous ginger",
        "forest dweller"
    ],
    "japanese anemone": [
        "tall asiatic perennial",
        "serene woodland flower",
        "east-asian native flower",
        "stoloniferous growth",
        "silky pink petal radiance",
        "woodland garden favorite",
        "east-asian perennial blossom",
        "late-season floral display",
        "stately forest edge flower",
        "tranquil late bloomer",
        "carpel rich center",
        "long-flowering autumnal star",
        "tall asiatic ornamental beauty",
        "delicate purplish-pink blossom array",
        "windflower delicate beauty"
    ],
    "black-eyed susan": [
        "rugged hairy rudbeckia",
        "coneflower-like black-eyed plant",
        "central black cone flower",
        "pollinator-friendly rudbeckia",
        "golden-yellow wildflower",
        "distinct black-eyed blossom",
        "sunny field susan",
        "daisy family rudbeckia",
        "showpiece planting",
        "fuzzy-stemmed rudbeckia",
        "brown-domed center",
        "golden petal surround",
        "easy-growing rudbeckia species",
        "sunny open site favorite",
        "bright American wildflower"
    ]
    ...
}
\end{lstlisting}
\begin{lstlisting}[basicstyle=\fontfamily{\ttdefault}\scriptsize, breaklines=true,caption={\textbf{Examples of generated descriptions for SUN.}}]
{
    "abbey": [
        "religious architecture",
        "peaceful prayer courtyard",
        "historic ecclesiastical complex",
        "monastic buildings",
        "carved stone reliefs",
        "silent contemplative gardens",
        "clerical chambers",
        "ancient liturgical hall",
        "sequestered religious refuge",
        "spiritual monastic center",
        "high-altitude abbey setting",
        "medieval landmark",
        "picturesque monastery",
        "tranquil cloister enclosure",
        "secluded pilgrimage destination",
        "ancestral cloister",
        "spirituality center",
        "parochial complex",
        "ecclesiastical estate",
        "tranquil abbey gardens",
        "monastic cellar vaults",
        "placid cloister quarters",
        "benedictine architecture",
        "clerical stonework",
        "Gregorian chant resonance",
        "secluded spiritual refuge",
        "illuminated manuscript repository",
        "divine service chapel",
        "clerestory window designs",
        "ancestral cloister"
    ],
    "airplane cabin": [
        "aeronautical window shape",
        "fuselage cross-section",
        "overhead bin latch",
        "cabin window shade",
        "recessed cabin lighting",
        "seatback tray table",
        "emergency exit handle",
        "aircraft cabin",
        "cabin crew interphone",
        "cabin altitude sign",
        "air nozzle",
        "avionics panel",
        "in-flight magazine pouch",
        "fuselage interior design",
        "cabin class divider"
    ],
    "airport terminal": [
        "terminal retail stores",
        "digital flight board",
        "entrance to arrivals",
        "airport lounge",
        "tax-free goods shop",
        "automated check-in spot",
        "security screening zone",
        "airport seating arrangements",
        "passenger check-in desks",
        "departure lounge seating",
        "airport check-in island",
        "gate podiums",
        "flight information boards"
    ],
    "engine room": [
        "turbine vibration control zone",
        "mechanical performance monitoring space",
        "heavy machinery operational platform",
        "maritime technical engineering station",
        "powertrain thermal regulation sector",
        "central power distribution chamber",
        "diesel generator maintenance location",
        "kinetic propulsion system area",
        "ship engine operation center",
        "marine engine service station",
        "mechanical room vibration analysis",
        "power distribution control centre",
        "engine diagnostic and repair shop",
        "power regeneration equipment area"
    ],
    "indoor escalator": [
        "dynamic stairway",
        "motorized belt",
        "visible step cleats",
        "longitudinal motion path",
        "smooth balustrade glass",
        "public indoor traversal",
        "automated staircase",
        "shopping center feature",
        "kinetic stair mechanism",
        "dynamic railing",
        "continuous movement",
        "moving handrail",
        "glass side panels",
        "flat escalator landing"
    ],
    "excavation": [
        "archaeological dig layers",
        "artefact excavation pits",
        "ancient civilization studies",
        "soil strata analysis",
        "dig site grid system",
        "stratigraphy profiles",
        "excavation site documentation",
        "earth sifting screens",
        "fossil dig uncovering",
        "sedimentary layer analysis",
        "archeological site",
        "dig zone",
        "soil screening",
        "historical excavation",
        "artifact discovery",
        "stratigraphy",
        "archeological survey",
        "relic preservation"
    ],
    "fairway": [
        "expansive fairway views",
        "tree-obstacle layout",
        "picturesque water features",
        "distinctive golf hole design",
        "strategic golf hole locations",
        "bunker-guarded green",
        "rough grass edges",
        "undulating golfer's terrain",
        "serpentine cart paths",
        "slope-contoured play lane"
    ]
    ...
}
\end{lstlisting}


\end{document}